\documentclass[letterpaper, 10 pt, conference]{ieeeconf}  

\IEEEoverridecommandlockouts                              

\usepackage{graphics} 
\usepackage{graphicx}
\usepackage{epsfig} 
\usepackage{amsmath} 
\usepackage{amssymb}  
\usepackage{subfigure}
\usepackage{float}
\usepackage{url}
\usepackage{multirow}
\usepackage{subfigure}
\usepackage{wrapfig}
\usepackage{tabularx}
\usepackage{tablefootnote}
\usepackage{hyperref}
\usepackage{color}
\usepackage{bm}
\usepackage{array}
\usepackage{booktabs}

\newcommand{\semi} {weakly}
\newcommand{\Semi} {Weakly}

\graphicspath{{./figs/}}

\newcolumntype{C}[1]{>{\centering\let\newline\\\arraybackslash\hspace{0pt}}m{#1}}

\newcommand{\doctitle}{Find Your Own Way: \Semi-Supervised Segmentation of Path Proposals for Urban Autonomy}
\newcommand{\docsubtitle}{}

\title{\LARGE \bf\doctitle\docsubtitle} 
\author{Dan Barnes, Will Maddern and Ingmar Posner 
\thanks{Authors are from the Oxford Robotics Institute, Dept. Engineering Science, University of Oxford, UK. 
{\tt\small \{dbarnes,wm,ingmar\}@robots.ox.ac.uk}}
}

\begin{document}
\maketitle
\thispagestyle{empty}
\pagestyle{empty}

\begin{abstract}

We present a \semi-supervised approach to segmenting proposed drivable paths in images with the goal of autonomous driving in complex urban environments.
Using recorded routes from a data collection vehicle, our proposed method generates vast quantities of labelled images containing proposed paths and obstacles without requiring manual annotation, which we then use to train a deep semantic segmentation network.  
With the trained network we can segment proposed paths and obstacles at run-time using a vehicle equipped with only a monocular camera without relying on explicit modelling of road or lane markings.
We evaluate our method on the large-scale KITTI and Oxford RobotCar datasets and demonstrate reliable path proposal and obstacle segmentation in a wide variety of environments under a range of lighting, weather and traffic conditions.
We illustrate how the method can generalise to multiple path proposals at intersections and outline plans to incorporate the system into a framework for autonomous urban driving. 

\end{abstract}

\vspace{-1mm}
\section{Introduction}\label{sec:introduction}

Road scene understanding is a critical component for decision making and safe operation of autonomous vehicles in urban environments.
Given the structured nature of on-road driving, all autonomous vehicles must follow the `rules of the road'; crucially, driving within designated lanes in the correct direction and negotiating intersections. 

Current commercial systems that perform driver assistance and on-road autonomy typically depend on visual recognition of lane markings and explicit definitions of lanes and traffic rules, and therefore rely on simple road layouts with clear markings (e.g. well-maintained highways) \cite{yenikaya2013keeping, hillel2014recent}.
To extend these systems beyond multi-lane highways to complex urban environments and rural or undeveloped locations without clear or consistent lane markings, an alternative approach is required.

\begin{figure}[t]
  \centering
  \includegraphics[width=1.0\columnwidth]{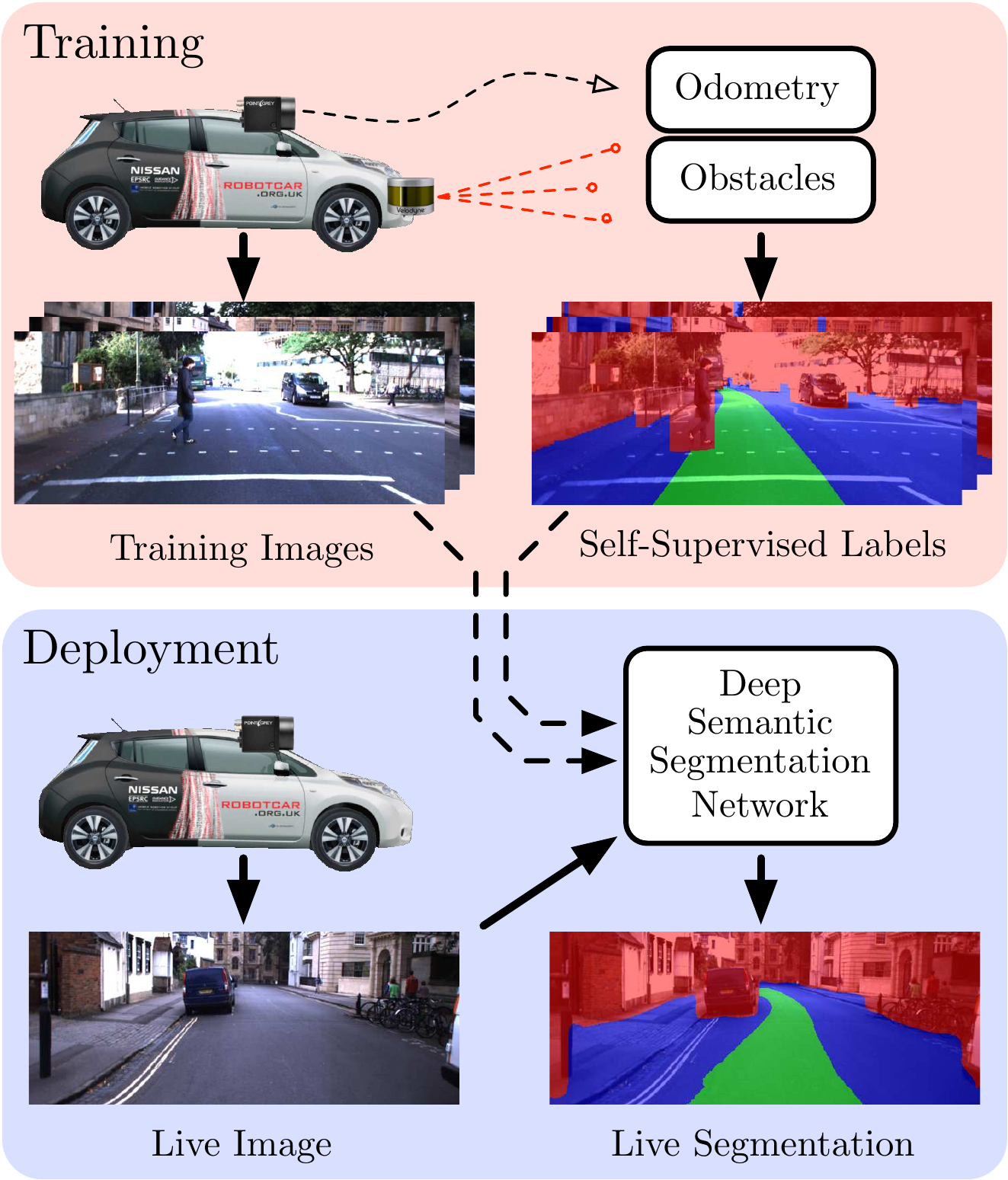}
  \caption{\Semi-supervised path proposal segmentation using our approach. A data collection vehicle equipped with a camera as well as odometry and obstacle sensors is used to collect vast quantities of data during normal driving (top). The odometry and obstacle data is projected into the training images to generate \semi-supervised labels relevant for on-road autonomy, which are then used to train a deep semantic segmentation network. At run-time, a vehicle equipped with only a monocular camera can perform live segmentation of the drivable path and obstacles using the trained network (bottom), even in the absence of explicit lane markings.}
  \label{fig:intro-figure}
  \vspace{-4mm}
  \vspace{-2mm}
\end{figure}

In this paper we present a \semi-supervised approach to segmenting \textit{path proposals} for a road vehicle in urban environments given a single monocular input image.
Our approach is capable of segmenting the proposed path for a vehicle in a diverse range of road scenes, without relying on explicit modelling of lanes or lane markings.
We define the term \textit{path proposal} as a route a driver would be expected to take through a particular road and traffic configuration.
We present a novel method of automatically generating labelled images containing path proposals.
Our method leverages both the behaviour of the data collection vehicle driver and additional sensors mounted to the vehicle, illustrated in Fig. \ref{fig:intro-figure} and the project video\footnotemark.
Using this approach we can generate vast quantities of labelled training data without any manual annotation, spanning a wide variety of road and traffic configurations under a number of different lighting and weather conditions limited only by the time spent driving the data collection vehicle.
We use this data to train an off-the-shelf deep semantic segmentation network (e.g. SegNet \cite{badrinarayanan2015segnet}) to produce path proposal segmentations using \emph{only} a monocular input image.

\setlength{\skip\footins}{.1cm}
\footnotetext{\color{blue}{\url{https://youtu.be/rbZ8ck_1nZk}}}
We evaluate our approach using two large-scale autonomous driving datasets: the KITTI dataset \cite{geiger2013vision}, collected in Karlsruhe, Germany, and the large-scale Oxford RobotCar Dataset \cite{RobotCarDatasetIJRR}, consisting of over 1000km of recorded driving in Oxford, UK, over the period of a year. 
For each of these datasets we make use of the additional sensors on the vehicle and the trajectory taken by the driver as the \semi-supervised signal to train a pixelwise semantic classifier.
We present segmentation results on the KITTI Road \cite{fritsch2013new}, Object and Tracking benchmarks and investigate the performance under different lighting and weather conditions using the Oxford dataset.



\vspace{-2mm}
\section{Related Work}\label{sec:related-work}

Traditional methods of camera-based drivable path estimation for road vehicles involve preprocessing steps to remove shadow and exposure artefacts \cite{alvarez2007shadow, katramados2009real}, extraction of low-level road and lane features \cite{mccall2006video, yamaguchi2008road}, fitting road and lane models to feature detections \cite{labayrade2006reliable, huang2011probabilistic}, and temporal fusion of road and lane hypotheses between successive frames \cite{jiang2009new, sawano2006road}. 
While effective in well-maintained road environments, these approaches suffer in the presence of occlusions, shadows and changing lighting conditions, unstructured roads, and areas with few or no markings \cite{hillel2014recent}.
Robustness can be significantly increased by combining images with radar \cite{ma2000simultaneous} or LIDAR \cite{huang2009finding} but at an increased sensor cost.

More recently, advances in image processing using deep learning \cite{lecun2015deep} have led to impressive results on the related problem of \textit{semantic segmentation}, which aims to provide per-pixel labels of semantically meaningful objects for input images \cite{long2015fully, papandreou2015weakly, badrinarayanan2015segnet}.
Deep networks make use of the full image context to perform semantic labelling of road and lane markings, and hence are significantly more robust than previous feature-based methods \cite{badrinarayanan2015segnet}.
However, for automated driving these approaches depend on large-scale manually-annotated road scene datasets (notably CamVid \cite{brostow2009semantic} and Cityscapes \cite{cordts2016cityscapes}, consisting of 700 and 5,000 labelled frames respectively), for which the labels are time-consuming and expensive to produce.

The challenges in building large-scale labelled datasets has led some researchers to consider virtual environments, for which ground-truth semantic labels can be rendered in parallel with synthetic camera images. 
Methods using customised video game engines have been used to produce hundreds of thousands of synthetic images with corresponding ground truth labels \cite{ros2016synthia, richter2016playing}. 
While virtual environments allow large-scale generation of ground-truth semantic labels, they present two problems: firstly, rendering pipelines are typically optimised for speed and may not accurately reflect real-world images (both above approaches suggest rendered images are used only for augmenting real-world datasets and hence manual labelling is still necessary); secondly, the actions of the vehicle and all other agents in the virtual world must be pre-programmed and may not resemble real-world traffic scenarios. 
A recent method uses sparse 3D prior information to transfer labels to real-world 2D images \cite{xie2016semantic} but requires sophisticated 3D reconstructions and manual 3D annotations.

Some approaches have proposed bypassing segmentation entirely and learning a direct mapping from input images to vehicle behaviour \cite{pomerleau1989alvinn, muller2005off}.
These methods also use the driver of the data collection vehicle to generate the supervised labels for the network (e.g. steering angle) and have recently demonstrated impressive results in real-world driving tests \cite{bojarski2016end}, but it is not clear how this approach generalises to scenarios where there are multiple possible drivable paths to consider (e.g. intersections). 
Our approach instead uses the data collection vehicle driver to implicitly label proposed paths in the image, but still allows a planning algorithm to choose the best path for the current route.



\section{\Semi-Supervised Segmentation}\label{sec:method}

In the following section we outline our approach for generating \semi-supervised training data for proposed path segmentation using video and sensor data recorded from a manually-driven vehicle.

\vspace{-1mm}
\subsection{Sensor Configuration}

In addition to a monocular camera to collect input images, our approach depends on the following two capabilities for the data collection vehicle:

\vspace{1mm}
\noindent
\textbf{Vehicle odometry:} a method of estimating the motion of the vehicle is required. For this we use stereo visual odometry \cite{WinstonChurchill}, although other methods using inertial systems or wheel odometry would suffice.
\vspace{1mm}

\noindent
\textbf{Obstacle sensing:} a method of detecting the 3D positions of impassible objects (both static and dynamic) in front of the vehicle is necessary to ensure that dynamic objects are not accidentally included in the drivable label area. For this we use a LIDAR scanner, though other methods that use dense stereo \cite{pfeiffer2010efficient} or automotive radar would also be suitable.

\vspace{1mm}

\begin{figure}[t]
    \centering
    \includegraphics[width=1.0\columnwidth]{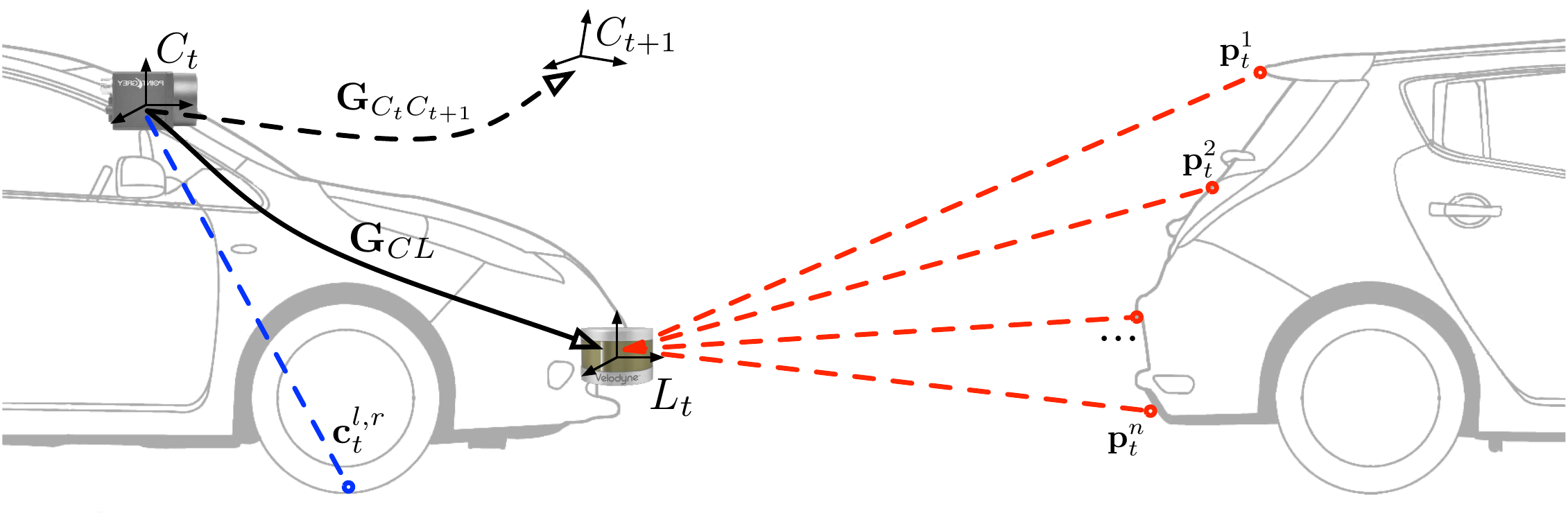}
    \vspace{-8mm} 
    \caption{Sensor extrinsics for \semi-supervised labelling. The survey vehicle (left) is equipped with a camera $C$ and obstacle sensor $L$, e.g. a LIDAR scanner. The extrinsic transform $\mathbf{G}_{CL}$ between the camera and LIDAR is found using a calibration routine. The contact point $\mathbf{c}^{\{l,r\}}$ of the left and right wheels on the ground relative to the camera frame $C$ is also measured at calibration time. At time $t$, the LIDAR scanner observes a number of points $\mathbf{p}_{t}^{1...n}$ on obstacles, including other vehicles on the road (right). The relative pose $\mathbf{G}_{C_{t}C_{t+1}}$ of the camera between time $t$ and $t+1$ is determined using vehicle odometry, e.g. using a stereo camera.}
    \label{fig:segmentation-diagram}
    \vspace{-6mm}
\end{figure}

Note that these additional sensing capabilities are only required for collecting training data; the resulting network only requires a monocular input image. Fig. \ref{fig:segmentation-diagram} illustrates the sensor extrinsics for a vehicle equipped with a stereo camera and LIDAR sensor.

\vspace{-1mm}
\subsection{\Semi-Supervised Labelling}

To generate class labels for pixels in the input image, we make use of large quantities of recorded data from the data collection vehicle driven by a human driver in a variety of traffic and weather conditions. 
We follow the general approach of methods that learn to drive by demonstration \cite{argall2009survey, silver2013learning}, and assume the \textit{proposed path} corresponds to the one chosen by the driver of the data collection vehicle in each scenario.
Labels are then generated by projecting the \textit{future} path of the vehicle into each image, over which object labels as detected by the LIDAR scanner are superimposed:

\subsubsection{Proposed path projection}

To project the future path of the vehicle into the current frame, it is necessary to know the size of the vehicle and the points of contact with the ground during the trajectory.
We assume the position of the contact points $\mathbf{c}_{\{l,r\}}$ of the front left and right wheels on the ground relative to the camera $C$ is determined as part of a calibration procedure.
The position of the contact point $\mathbf{c}_{\{l,r\}}$ in the current camera frame $C_{t}$ after $k$ frames is then found as follows: 

\vspace{-2mm}
\begin{equation}\label{eqn:camera-vo-projection}
  ^{C_{t}}\mathbf{c}_{\{l,r\},k}=\mathbf{K}\mathbf{G}_{C_{t}C_{t+k}}\mathbf{c}_{\{l,r\}}
\end{equation}

\noindent
where $\mathbf{K}$ is the perspective projection matrix for the camera $C$ and $\mathbf{G}_{C_{t}C_{t+k}}$ is the $\mathbb{SE}(3)$ chain of relative pose transforms formed by vehicle odometry from frame $t$ to frame $t+k$ as follows:

\vspace{-1mm}

\vspace{-1mm}
\vspace{-2mm}
\begin{equation}\label{eqn:vo-transform-chain}
  \mathbf{G}_{C_{t}C_{t+k}} = \mathbf{G}_{C_{t}C_{t+1}}\times\mathbf{G}_{C_{t+1}C_{t+2}}\times\dots\times\mathbf{G}_{C_{t+k-1}C_{t+k}}
\end{equation}

\begin{figure}[t]
  \centering
  \includegraphics[width=1.0\columnwidth]{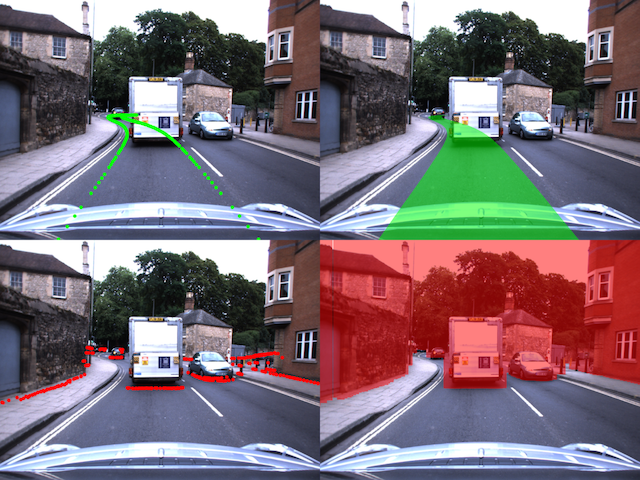}
  \caption{Ground contact point (top) and obstacle point (bottom) projection into images. At time $t$, ground contact points $\mathbf{c}_{\{l,r\},j}$ (green) corresponding to the path of the vehicle up to $k$ frames ahead are projected into the current image (top left). Pixel labels corresponding to drivable paths are filled in by drawing quadrilaterals between the left and right contact points between two successive frames (top right). At the same time, obstacle points $\mathbf{p}_{t}^{i}$ (red) from the current LIDAR scan are projected into the image (bottom left). Pixel labels corresponding to obstacles are formed by extending each of these points to the top of the image (bottom right). Note that the top and bottom sections of the image corresponding to the sky and vehicle bonnet are removed before training.}
  \label{fig:vo-and-obstacles-to-mask}
  \vspace{-4mm}
\end{figure}

Proposed path pixel labels are then formed by filling quadrilaterals in image coordinates corresponding to sequential future frames.
The vertices of the quadrilateral are formed by the following points in camera frame $C_{t}$:

\vspace{-2mm}
\begin{equation}\label{eqn:path-vertices}
  \left\{^{C_{t}}\mathbf{c}_{l,j},^{C_{t}}\mathbf{c}_{l,j-1},^{C_{t}}\mathbf{c}_{r,j-1},^{C_{t}}\mathbf{c}_{r,j}\right\}
\end{equation}

\noindent
where the index variable $j=\{1\dots k\}$. 
An illustration of the proposed path projection and labelling process is shown in Fig. \ref{fig:vo-and-obstacles-to-mask}.
The choice of frame count $k$ depends on the look-ahead distance required for path labelling and the accuracy of the vehicle odometry system used to provide relative frame transforms.
In practice we choose $k$ such that the distance between first and last contact points $\left\|\mathbf{G}_{C_{t}C_{t+k}}\mathbf{c}_{\{l,r\}} - \mathbf{c}_{\{l,r\}}\right\|$ exceeds 60 metres.
Different camera setups with higher viewpoints may require greater path distances, but accumulated odometry error will affect far-field projections.

\subsubsection{Obstacle projection}\label{sec:obstacle-projection}

\begin{figure*}[t]
  \centering
  \includegraphics[width=1.0\textwidth]{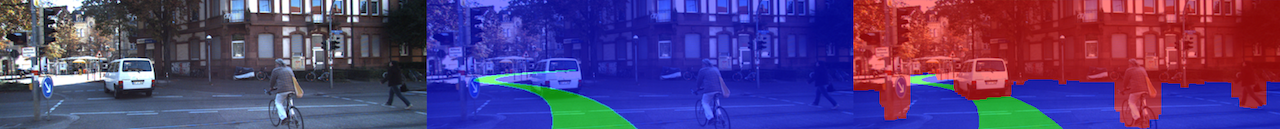}
  \caption{Proposed path labels for an input image (left) before (middle) and after (right) applying obstacle labels from the LIDAR scanner. Without the obstacle labels, the proposed path (middle, green) intersects vehicles in the same lane as the path driven by the data collection vehicle, which in this case will erroneously label sections of the white van as drivable route. Adding labels for obstacles (right, red) ensures that dynamic objects including the van, cyclist and pedestrian are marked as non-drivable. Note that static obstacles such as the road sign and the building are also labelled as obstacles, which correctly handles occlusions (e.g. as the path turns right after the traffic lights).}
  \label{fig:labels-before-after-obstacles}
  \vspace{-4mm}
\end{figure*}

For some applications it may be sufficient to use just the proposed path labels to train a semantic segmentation network.
However, for on-road applications in the presence of other vehicles and dynamic objects, a naive projection of the path driven will intersect vehicles in the same lane and label them as drivable paths as illustrated in Fig. \ref{fig:labels-before-after-obstacles}.
This may lead to catastrophic results when the labelled images are used to plan paths for autonomous driving, since vehicles and traffic may be labelled as traversable by the network.

We make use of the obstacle sensor mounted on the vehicle, in our case a LIDAR scanner. 
Each 3D obstacle point $\mathbf{p}_{t}^{i}$ observed at time $t$ is projected into the camera frame $C_{t}$ as follows:

\vspace{-1mm}
\vspace{-2mm}
\begin{equation}\label{eqn:lidar-obstacle-projection}
  ^{C_{t}}\mathbf{p}_{t}^{i}=\mathbf{K}\mathbf{G}_{CL}\mathbf{p}_{t}^{i}
\end{equation}

\noindent
where $\mathbf{K}$ is the camera projection matrix and $\mathbf{G}_{CL}$ is the $\mathbb{SE}(3)$ extrinsic calibration between the camera and LIDAR sensor.
For each camera-frame point $^{C_{t}}\mathbf{p}_{t}^{i}$, we take an approach inspired by ``stixels'' \cite{pfeiffer2010efficient, scharwachter2014stixmantics} and label all pixels in the image on and above the point as an obstacle.
This ensures all locations above and behind the detected obstacle are labelled as non-drivable, as illustrated in Fig. \ref{fig:vo-and-obstacles-to-mask}.
Obstacle pixel labels take precedence over proposed path labels to ensure correct labelling of safe drivable paths as illustrated in Fig. \ref{fig:labels-before-after-obstacles}.

In most images, there will be locations labelled as neither proposed path nor obstacle.
These correspond to locations which the vehicle has not traversed, and no positive identification of obstacles have been made. 
Typically these areas correspond to the road area outside the current lane (including lanes for oncoming traffic), kerbs, empty pavements and ditches.
We refer to these locations as ``unknown area'' since it is not clear whether the vehicle should enter these spaces; this would be a decision for a higher-level planning framework as discussed in Section \ref{sec:conclusions}.


\vspace{-1mm}
\subsection{Semantic Segmentation}\label{sec:semantic-segmentation}

Once proposed path, obstacle and unknown area labels are automatically generated for a large number of recorded images, they can be used to train a semantic segmentation network to classify new images from a different vehicle equipped with only a monocular camera.
We make use of SegNet \cite{badrinarayanan2015segnet}, a deep convolutional encoder-decoder architecture for pixelwise semantic segmentation.
Although higher-performing network architectures now exist (e.g. \cite{papandreou2015weakly}), SegNet provides real-time evaluation on consumer GPUs, making it suitable for deployment in an autonomous vehicle.

\begin{figure}[b]
    \vspace{-3mm}
    \centering
    \includegraphics[width=1.0\columnwidth]{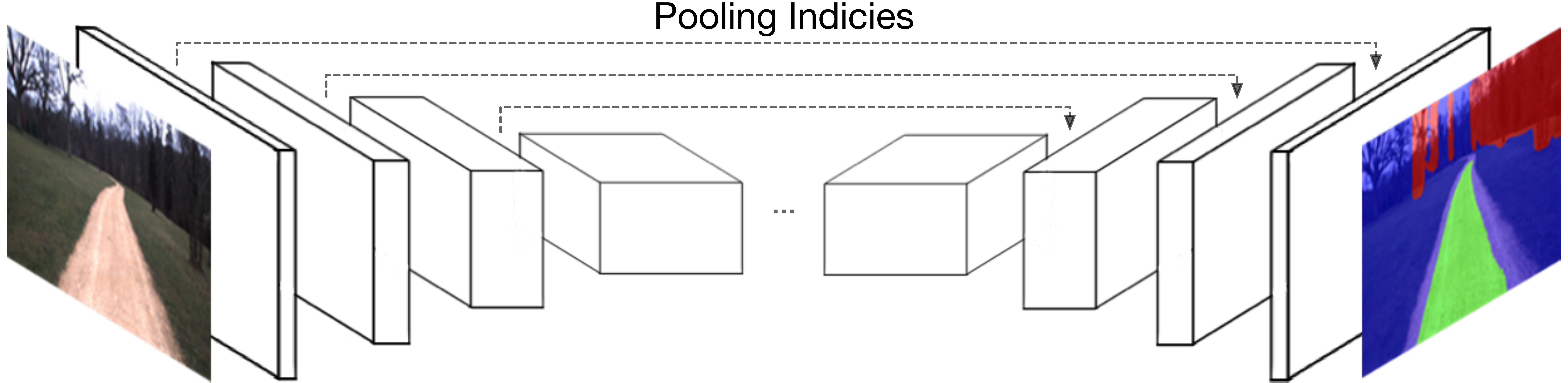}
    \caption{Semantic segmentation is performed using a common encoder-decoder architecture (e.g. \cite{badrinarayanan2015segnet}) where the feature representation is progressively spatially compressed before being expanded to a full resolution per-pixel class prediction.}
    \label{fig:network-example}
\end{figure}

The \semi-supervised labelling approach described in this section can generate vast quantities of training data, limited only by the length of time spent driving the data collection vehicle.
However, the types of routes driven will also bias the input data, as most on-road driving is performed in a straight line; a random subsample of the training data will consist mostly of straight-line driving.
In practice we subsample the data to 4Hz, before further subsampling based on turning angle.
For each frame we compute the average yaw rate $\bar{\Delta\psi}$ per frame for the corresponding proposed path as follows:
\vspace{-2mm}
\begin{equation}\label{eqn:turning-angle}
  \bar{\Delta\psi}=\dfrac{1}{k}\sum_{i}^{k}\psi\left(\mathbf{G}_{C_{t+i-1}C_{t+i}}\right)
\end{equation}

\noindent
where $\psi(\mathbf{G})$ is a function that extracts the Euler yaw angle $\psi$ from the $\mathbb{SE}(3)$ transform matrix $\mathbf{G}$. 
We then build a histogram of average yaw rates and randomly sample from the histogram bins to ensure an unbiased selection of different turning angles. 



\section{Experimental Setup}\label{sec:experimental-setup}

We build two different models for evaluation: one using the KITTI Raw dataset \cite{geiger2013vision} and one using the Oxford RobotCar dataset.
These datasets were collected using different vehicles with different sensor setups, summarised in Table \ref{tab:training-setup}.

\begin{table}
\centering
\caption{Vehicle and setup summary}
\vspace{-1mm}
\begin{tabular}{C{1.4cm} | C{3.3cm} | C{2.5cm} }
{\bf Vehicle} & {\parbox{2.6cm}{\centering \bf Oxford RobotCar Nissan LEAF}} & {\bf KIT AnnieWAY VW Passat} \\ \hline
{\bf Camera Sensor}
    & Point Grey Bumblebee XB3
    & 2 x Point Grey Flea2 \\ \hline
{\bf Input Resolution} 
    & 640 x 256
    & 621 x 187 \\ \hline
{\bf LIDAR} 
    & {\parbox{2.6cm}{\centering SICK LD-MRS 4-beam}}  
    & Velodyne HDL-64E 64-beam \\ \hline
{\bf Vehicle Width} 
    & 2.43 m 
    & 2.2 m
\end{tabular}
\label{tab:training-setup}
\vspace{-5mm}
\end{table}

\subsection{Platform Specifications}

\begin{figure*}[t]
    \centering
    \includegraphics[width=1.0\textwidth]{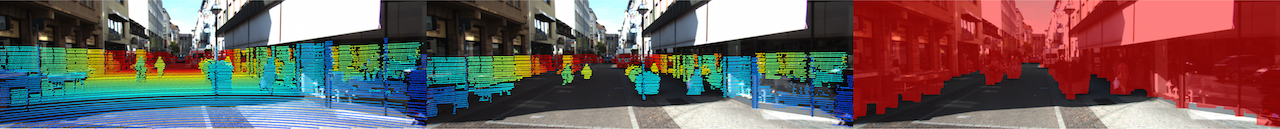}
    \caption{Obstacle labelling using Velodyne data for the KITTI dataset. Raw Velodyne scans (left) contain returns from the road surface as well as nearby obstacles. We fit a ground plane using MLESAC and retain only points 0.25m above the plane (middle). We then label pixels using the approach in Section \ref{sec:obstacle-projection} (right) to ensure accurate labels on obstacles while retaining drivable surfaces on the ground.}
    \label{fig:velodyne-plane-fit}
    \vspace{-4mm}
\end{figure*}

Both vehicles are equipped with stereo camera systems, and we use the stereo visual odometry approach in \cite{WinstonChurchill} to compute the relative motion estimates required in Eq. \ref{eqn:vo-transform-chain}.
The images from the cameras are cropped and downscaled to the resolutions listed in Table \ref{tab:training-setup} before training.
The Oxford RobotCar is equipped with a SICK LD-MRS LIDAR scanner, which performs obstacle merging and tracking across 4 scanning planes in hardware.
We use points identified as ``object contours'' to remove erroneous obstacles due to noise and ground-strike.
The Velodyne HDL-64E mounted on the AnnieWAY vehicle does not perform any object filtering, and hence we use the following approach to detect obstacles:
we fit a ground plane to the 3D LIDAR scan using MLESAC \cite{torr2000mlesac}, and treat all points more than 0.25m above this plane as obstacles, as illustrated in Fig. \ref{fig:velodyne-plane-fit}.
This approach effectively identifies obstacles the vehicle may collide with even in the presence of pitching and rolling motions.
The camera-LIDAR calibration $\mathbf{G}_{CL}$ for the RobotCar vehicle was determined using the method in \cite{pascoe2015direct}; for the AnnieWAY vehicle the calibration provided with the KITTI Raw dataset was used.

\vspace{-1mm}

\subsection{Network Training}\label{sec:network-training}

\begin{figure}[t]
  \centering
  \includegraphics[width=1.0\columnwidth]{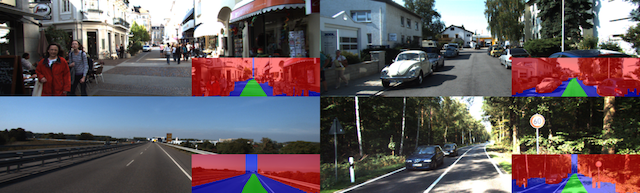}
  \includegraphics[width=1.0\columnwidth]{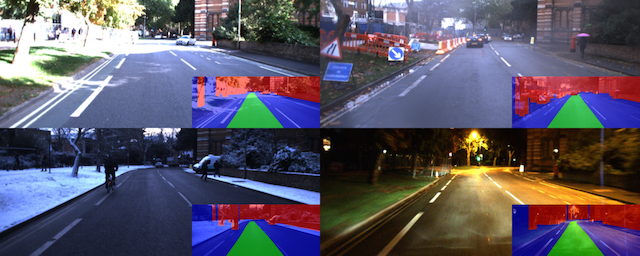}
  \caption{Example training images with \semi-supervised labels from the KITTI (top) and Oxford (bottom) datasets. The \semi-supervised approach generates proposed path and obstacle labels for a diverse set of locations in the KITTI dataset, and a diverse set of conditions for the same location in the Oxford dataset. No manual annotation is required to generate the labels.}
  \label{fig:TrainingConditions}
  \vspace{-5mm}
\end{figure}

For the KITTI model, we made use of the available City, Residential and Road data from the KITTI Raw dataset.
For the Oxford model, we selected a diverse range of weather conditions for each traversal of the route, including 9 overcast, 8 with direct sun, 4 with rain, 2 at night and 1 with snow; each traversal consisted of approximately 10km of driving.
The number of labelled images used to train each model is shown in Table \ref{tab:training-data-statistics} and some examples are shown in Fig. \ref{fig:TrainingConditions}.
In total we used 24,443 images to train the KITTI model, and 49,980 images for the Oxford model.

\setlength{\skip\footins}{0.3cm}

\begin{table}
\centering
\caption{Training image summary statistics}
\begin{tabular}{C{2cm} | C{2cm} | C{2cm} }
{\bf Dataset} & {\bf Condition} & {\bf Training Images} \\ \hline
\multirow{4}{*}{KITTI\tablefootnote{\color{blue}{\url{http://www.cvlibs.net/datasets/kitti/}}}}
& City          & 1264  \\ 
& Residential   & 20734 \\ 
& Road          & 2445  \\ 
& \bf{Total}         & \bf{24443} \\ \hline
\multirow{6}{*}{Oxford\tablefootnote{\color{blue}{\url{http://robotcar-dataset.robots.ox.ac.uk}}}}
& Overcast      & 17085  \\ 
& Sun           & 16299 \\ 
& Rain          & 9822 \\ 
& Night         & 4170 \\ 
& Snow          & 2604 \\ 
& \bf{Total}         & \bf{49980}
\end{tabular}
\label{tab:training-data-statistics}
\vspace{-4mm}
\end{table}

For both datasets we built semantic classifier models using the standard SegNet convolutional encoder-decoder architecture.
The same SegNet parameters were used for both datasets, with modifications only to account for the differences in input image resolution.
We randomly split the input data into 75\% training and 25\% validation sets, performed training for 100 epochs then selected the best-performing model according to the validation set results.

For the comparison using the KITTI Road benchmark presented in Section \ref{sec:kitti-road-evaluation}, we trained an additional SegNet model on only the training images provided for the Ego-Lane Estimation Evaluation.
Note that these ground truth images were not provided to the model trained using the \semi-supervised approach described above.
For the object detection evaluation using the KITTI Object and Tracking datasets, we have ensured that there is no overlap between images selected to train the \semi-supervised labels and the images with ground truth labels used in the evaluation.



%


\section{Results}\label{sec:results}

For reliable on-road driving, the semantic segmentation must function in multiple environments under the range of lighting, weather and traffic conditions encountered during normal operation.
In this section we evaluate the performance of both the KITTI model and Oxford model under a range of different test conditions.

\subsection{Oxford Dataset}

\begin{figure}[t]
  \centering
  \includegraphics[width=1.0\columnwidth]{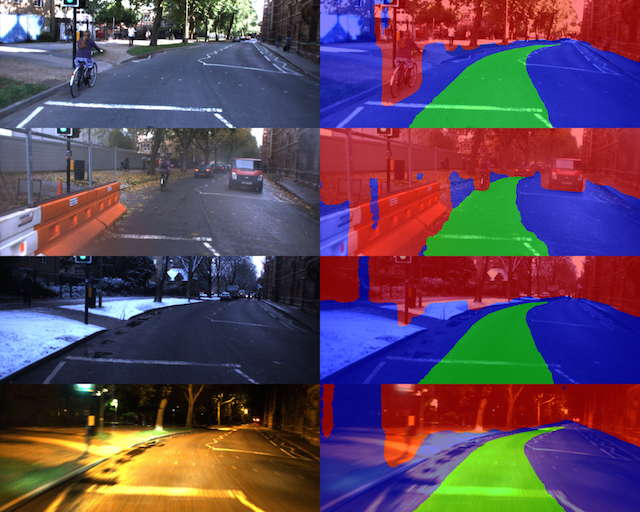}
  \caption{Semantic segmentation on frames captured at the same location under different conditions. Despite significant changes in appearance between sunny, rainy, snowy and night-time conditions, the network correctly segments the proposed drivable path and labels obstacles including cyclists, other vehicles and road barriers.}
  \label{fig:SegmentationUnderDifferentConditions}
  \vspace{-2mm}
  \vspace{-4mm}
\end{figure}

\begin{figure}[]
  \centering
  \includegraphics[width=1.0\columnwidth]{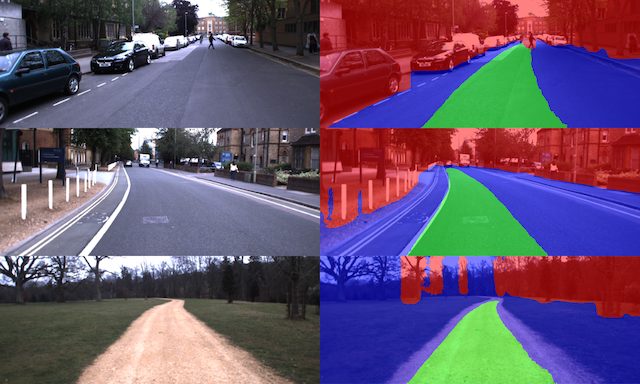}
  \caption{Path proposals in locations without explicit lane dividers or road markings. Using the context of the road scene the network infers the correct proposed path (top, middle), even for gravel roads never seen in the training data (bottom).}
  \label{fig:no-road-markings}
  \vspace{-2mm}
  \vspace{-3mm}
\end{figure}

We evaluate the Oxford model by generating ground truth labels for a further four datasets not used for training, consisting of 2,718 images in sunny conditions, 2,481 images in cloudy conditions, 2,340 images collected at night and 1,821 images collected in the rain, for a total of 9,360 test images. 
Table \ref{tab:MRGSegmentationResults} presents the segmentation results for the three classes in each of the four different conditions in the test datasets listed above, where the ``All'' column shows the mean of precision (PRE), recall (REC) and intersection-over-union (IoU) across all classes.
The model provides good performance across the different conditions with mean IoU scores exceeding 80\% in all cases, with the highest performance in cloudy weather and lowest at night, due to the reduced image quality in low-light conditions.
Fig. \ref{fig:SegmentationUnderDifferentConditions} illustrates the output of the network for four images of the same location under different conditions. Despite significant changes in lighting and weather, the network correctly determines the proposed path through the crossing and identifies obstacles (e.g. construction barriers).
This result demonstrates that the \semi-supervised approach can be used to train a single network that segments proposed paths and obstacles across a wide range of conditions without explicitly modelling environmental changes due to lighting, weather and traffic.
Fig. \ref{fig:no-road-markings} presents a number of locations where the network proposed a valid path in the absence of explicit road or lane markings, instead using the context of the road scene to infer the correct route. 
\vspace{-2mm}

\begin{table*}
\centering
\caption{Segmentation results for Oxford test data across varying conditions.}
\begin{tabular}{C{2cm} | C{3cm} | C{3cm} | C{3cm} | C{3cm} }
{\bf Condition}& {\bf Proposed Path} & {\bf Obstacle} & {\bf Unknown Area}  & {\bf All} \\ \hline
Night       
& \begin{tabular}{r l} PRE & 86.50\% \\ REC & 87.75\% \\ IoU & 77.18\% \\ \end{tabular}
& \begin{tabular}{r l} PRE & 93.60\% \\ REC & 93.71\% \\ IoU & 88.06\% \\ \end{tabular}
& \begin{tabular}{r l} PRE & 88.88\% \\ REC & 88.31\% \\ IoU & 79.53\% \\ \end{tabular}
& \begin{tabular}{r l} PRE & 89.66\% \\ REC & 89.92\% \\ IoU & 81.59\% \\ \end{tabular}
\\ \hline
Rain       
& \begin{tabular}{r l} PRE & 89.55\% \\ REC & 86.97\% \\ IoU & 78.95\% \\ \end{tabular}
& \begin{tabular}{r l} PRE & 94.04\% \\ REC & 96.88\% \\ IoU & 91.27\% \\ \end{tabular}
& \begin{tabular}{r l} PRE & 91.41\% \\ REC & 88.73\% \\ IoU & 81.90\% \\ \end{tabular}
& \begin{tabular}{r l} PRE & 91.66\% \\ REC & 90.86\% \\ IoU & 84.04\% \\ \end{tabular}
\\ \hline
Overcast       
& {\bf \begin{tabular}{r l} PRE & 91.13\% \\ REC & 92.63\% \\ IoU & 84.97\% \\ \end{tabular}}
& \begin{tabular}{r l} PRE & 94.76\% \\ REC & 96.68\% \\ IoU & 91.77\% \\ \end{tabular}
& {\bf \begin{tabular}{r l} PRE & 93.41\% \\ REC & 90.53\% \\ IoU & 85.09\% \\ \end{tabular}}
& {\bf \begin{tabular}{r l} PRE & 93.10\% \\ REC & 93.28\% \\ IoU & 87.27\% \\ \end{tabular}}
\\ \hline
Sun       
& \begin{tabular}{r l} PRE & 89.50\% \\ REC & 89.53\% \\ IoU & 81.02\% \\ \end{tabular}
& {\bf \begin{tabular}{r l} PRE & 94.85\% \\ REC & 97.01\% \\ IoU & 92.16\% \\ \end{tabular}}
& \begin{tabular}{r l} PRE & 92.56\% \\ REC & 90.05\% \\ IoU & 83.97\% \\ \end{tabular}
& \begin{tabular}{r l} PRE & 92.30\% \\ REC & 92.20\% \\ IoU & 85.72\% \\ \end{tabular}
\end{tabular}
\label{tab:MRGSegmentationResults}
\vspace{-3mm}
\end{table*}

\subsection{KITTI Benchmarks}


To demonstrate how our \semi-supervised labelling approach can lead to useful performance for autonomous driving tasks, we evaluate it on two different benchmarks from the KITTI Vision Benchmark Suite: ego-lane segmentation and object detection.
However, neither of these benchmarks are an exact match for the segmentation results provided by the network, as they were designed for different purposes.
Accordingly, we present alternative metrics based on the provided ground truth to quantitatively evaluate our system. 
Note that the following two sections present different evaluation metrics for the same model trained on the same input data and should be interpreted in concert; the network produces both path and obstacle labels for each test image even when only one class is under evaluation.

\subsubsection{Ego-lane Segmentation}\label{sec:kitti-road-evaluation}

\begin{figure*}[]
  \centering
  \includegraphics[width=0.885\textwidth]{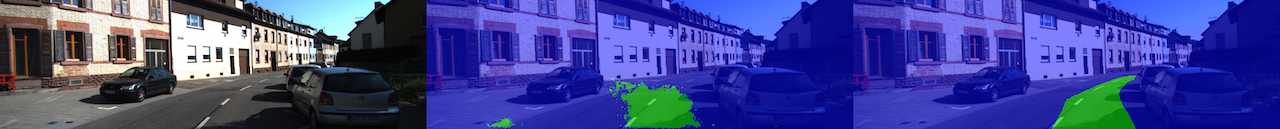}
  \caption{Example ego-lane segmentation results using the KITTI Road dataset. For the given input image (left), a SegNet model trained on the small number of manually-annotated ground truth images (middle) performs poorly in comparison with the model trained on the much larger \semi-supervised dataset (right) generated without manual annotation.}
  \label{fig:ego-lane-bl-vs-ss}
\end{figure*}

The closest analogue to a \textit{proposed path} in the KITTI benchmark suite is the ego-lane, consisting of the entire drivable surface within the lane the vehicle currently occupies \cite{fritsch2013new}.
The ego-lane dataset consists of 95 training and 96 test images, each with manually annotated ground truth labels.
We trained an additional SegNet model on the provided ground truth training images to compare to our model trained on \semi-supervised labelled images, as detailed in Section \ref{sec:network-training}.
The results of both models on the KITTI website benchmark is shown in Table \ref{tab:EgolaneSegmentationResults}.
Fig. \ref{fig:ego-lane-bl-vs-ss} illustrates a sample network output for both models.
The \semi-supervised model outperforms the model trained on the provided ground-truth images, with a 20\% increase in max F-score and 15\% increase in precision exceeding 90\% in total, despite never making use of manually annotated ground truth images or explicit encoding of lane markings.
Although the overall performance is not competitive with those generated by more sophisticated network architectures on the KITTI leaderboard (due to the different definition of ego-lane and proposed path), this result strongly indicates that the \semi-supervised approach generates segmentations useful for real-world path planning. 
The differences in the number of training images used for each model is illustrative of the fact that manually-annotated datasets will always be more be more time-consuming and expensive to produce than our \semi-supervised approach; even if manually annotated data is also available, for many tasks our approach could be used as pre-training to further improve results.

\begin{table*}[]
\centering
\caption{Ego-lane segmentation results on the KITTI Road benchmark}
\begin{tabular}{C{3cm} | C{2cm} | C{1.5cm} | C{1.5cm} | C{1.5cm} | C{1.5cm} | C{1.5cm} | C{1.5cm}}
{\bf Training} & {\bf Benchmark} & {\bf MaxF} & {\bf AP} & {\bf PRE} & {\bf REC} & {\bf FPR} & {\bf FNR} \\ \hline
\multirow{1}{*}{Provided}               & UM\_LANE  & 52.42\% & 37.85\% & 77.88\% & 39.50\% & 1.98\% & 60.50\%  \\ \hline
\multirow{1}{*}{\Semi-Supervised}  & UM\_LANE  & {\bf 72.88}\% & {\bf 64.49}\% & {\bf 92.78}\% & {\bf 60.01}\% & {\bf 0.82}\% & {\bf 39.99}\%
\end{tabular}
\label{tab:EgolaneSegmentationResults}
\end{table*}

\begin{figure*}[!]
  \centering
  \includegraphics[width=0.885\textwidth]{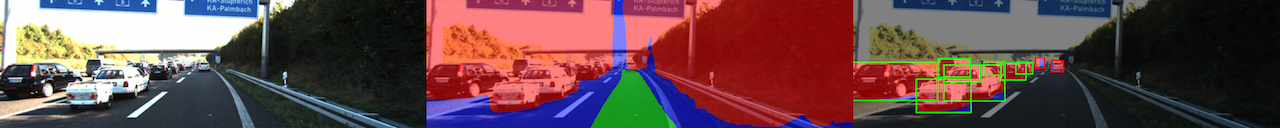}
  \caption{Example object detection results using obstacle segmentation. For a given input image (left), the network labels areas corresponding to proposed path, obstacle and unknown area (middle). For each ground-truth bounding box provided in the KITTI Object and Tracking datasets, we compute the ratio of pixels labelled as obstacle by our method (right). For each object instance, we consider it detected (green outline) if more than 75\% of the pixels within the bounding box are labelled as obstacles. Note that even for failed detections (red outline), a number of the pixels were still labelled as obstacle, and due to the tight obstacle outlines provided by our method we may miss portions of the bounding box (e.g. undercarriage of vehicles at bottom left).}
  \label{fig:obstacle-segmentation-example}
  \vspace{-3mm}
\end{figure*}

\subsubsection{Object Detection}

While the KITTI benchmark suite does not contain a semantic segmentation benchmark, it does contain object instance bounding boxes in both the Object and Tracking datasets. 
The definition of an \emph{object} in the KITTI benchmark (an individual instance of a vehicle or person within a bounding box) differs significantly from our definition of an \emph{obstacle} as part of the \semi-supervised approach (any part of the scene the vehicle might collide with).
However, we can evaluate object detection performance by ensuring that every \emph{object} instance provided by the KITTI Object and Tracking benchmarks was also classified as an \emph{obstacle} by our segmentation approach; hence we aim for the highest \emph{pixel-wise recall} score.
For each object instance we evaluate the number of pixels within the bounding box classified as an obstacle using our \semi-supervised approach, as illustrated in Fig. \ref{fig:obstacle-segmentation-example}.
We present three different recall metrics: pixel recall, which includes all pixels under all bounding boxes for each object class, and two variants of instance recall, which requires a certain fraction of obstacle-labelled pixels within each bounding box instance before the object is considered as ``detected'' (thresholds of 50\% and 75\% are presented). 
We present recall results on the data provided as part of the Object and Tracking datasets (consisting of 15,047 images with 87,343 total object instances) in Table \ref{tab:ObstacleSegmentationResults}, and an example detection is shown in Fig. \ref{fig:obstacle-segmentation-example}.
We have combined the object classes as follows: car, van, truck and tram labels are grouped as Vehicle; pedestrian, person sitting and cyclist labels are grouped as Person, and all others are grouped as Misc.
The results show that the \semi-supervised segmentation approach is reliably labelling \emph{objects} as \emph{obstacles} regardless of object class (and performs especially well for an instance recall threshold of 50\%); this is critical to avoid planning trajectories that intersect other vehicles or road users. 


\begin{table}[t]
\centering
\caption{Obstacle segmentation results on the KITTI Object and Tracking datasets}
\begin{tabular}{C{2.8cm} | C{0.9cm} | C{0.9cm} | C{0.9cm} | C{0.9cm}  }
{\bf Metric} & {\bf Vehicle} & {\bf Person} & {\bf Misc} & {\bf All} \\ \hline
Pixel Recall                            & 93.73\%     & 92.47\%     & 94.11\%     & 93.53\% \\ \hline
Instance Recall (\textgreater 50\%)     & 99.52\%     & 99.65\%     & 99.29\%     & 99.55\% \\ \hline
Instance Recall (\textgreater 75\%)     & 98.15\%     & 97.38\%     & 96.73\%     & 97.93\% \\ 
\end{tabular}
\label{tab:ObstacleSegmentationResults}
\end{table}
\subsection{Limitations}

\begin{figure}[]
  \centering
  \includegraphics[width=1.0\columnwidth]{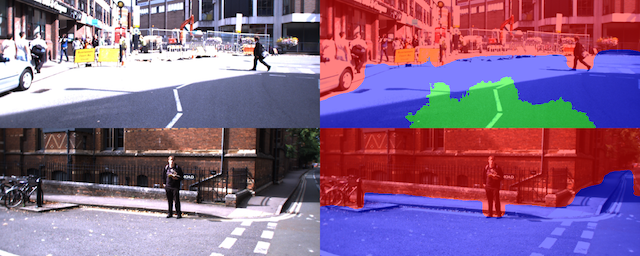}
  \caption{Proposed path segmentation failures. (Top) Overexposed or underexposed images will lead to incorrect path segmentation; this could be addressed by using a high-dynamic-range camera. (Bottom) At some intersections during tight turns, there is no clear path to segment as it falls outside the field of view of the camera; using a wider field of view lens or multiple cameras in a surround configuration would address this limitation.}
  \label{fig:limitations}
  \vspace{-4mm}
\end{figure}

Under some conditions the network fails to produce useful proposed path segmentations, as illustrated in Fig. \ref{fig:limitations}. 
These failure cases are mostly due to limitations of the sensor suite (poor exposure or low field of view), and could be addressed using a larger number of higher-quality cameras.

\subsection{Route Generalisation}

\begin{figure*}[t]
  \centering
  \includegraphics[width=0.845\textwidth]{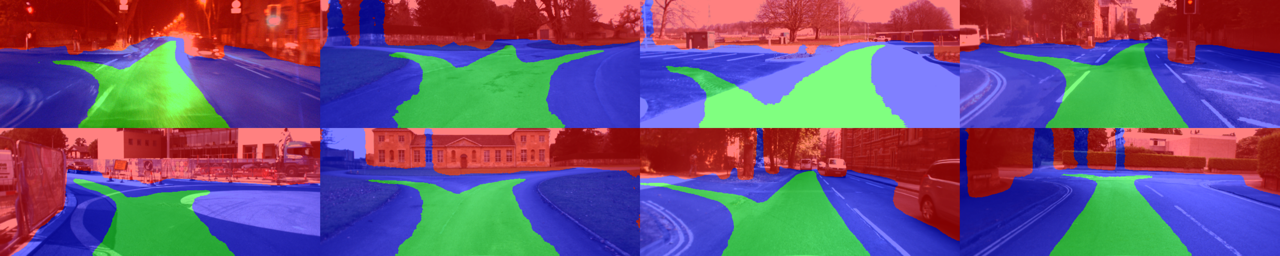}
  \caption{Proposed path generalisation to multiple routes. At intersections and roundabouts the network will often label different possible paths, which can then be leveraged by a planning framework for decision making during autonomous navigation.} 
  \label{fig:Generalisation}
  \vspace{-5.85mm}
\end{figure*}

As the \semi-supervised labels are generated from the recording of a data collection trajectory, it can only provide one proposed path per image at training time.
However, at intersections and other locations with multiple possible routes, at test time the resulting network frequently labels multiple possible proposed paths in the image as shown in Fig. \ref{fig:Generalisation}; this is an important step towards decision-making for topological navigation within a road network.
Currently we have no ground truth to evaluate route generalisation; we present qualitative results here for illustration only and plan to further characterise this effect in a future publication. 

\section{Conclusions}\label{sec:conclusions}

In this paper we have outlined our approach for \semi-supervised labelling of images for proposed path segmentation during on-road driving using only a monocular camera.
We have demonstrated that by leveraging multiple sensors and the behaviour of the data collection vehicle driver, we are able to generate vast quantities of semantically-labelled training data relevant for autonomous driving applications; crucially, we do not require any manual labelling of images in order to train our segmentation network.  
Our approach does not depend on specific road markings or explicit modelling of lanes to propose drivable paths. 
We evaluated the approach in the context of ego-lane segmentation and obstacle detection using the KITTI dataset, outperforming networks trained on manually-annotated training data and providing reliable obstacle detections.
We also demonstrated the robustness of the trained network to changes in lighting, weather and traffic conditions using the large-scale Oxford RobotCar dataset, with successful proposed path segmentation in sunny, cloudy, rainy, snowy and night-time conditions.
We plan to integrate the network with a planning framework that includes our previous work on topometric localisation across experiences \cite{linegar2016made} as well as our semantic map-guided approach for traffic light detection \cite{barnes2015exploiting} to enable fully autonomous driving in complex urban environments.

\section{Acknowledgements}\label{sec:conclusions}

This work was supported by the UK EPSRC Doctoral Training Partnership and Programme Grant EP/M019918/1.



\bibliographystyle{IEEEtran}
\bibliography{icra2017}

\end{document}